\documentclass[conference]{IEEEtran}
\PassOptionsToPackage{caption=false,font=footnotesize}{subfig}
\usepackage{subfig}  
\IEEEoverridecommandlockouts
\usepackage{cite}
\usepackage{amsmath,amssymb,amsfonts}

\usepackage{array}
\usepackage{scalerel}
\usepackage{multicol}
\usepackage{esdiff}

\usepackage{cases}
\usepackage{lipsum}
\usepackage{balance}
\usepackage{algorithm}
\usepackage{placeins}
\usepackage{dblfloatfix}
\usepackage{graphicx}
\usepackage{textcomp}
\usepackage{algpseudocode}

\usepackage{hyperref}
\usepackage{cleveref}
\usepackage{mathtools}
\usepackage{multirow}

\usepackage{booktabs}
\usepackage[table,dvipsnames]{xcolor} 
\usepackage{tabularx}
\usepackage{threeparttable}
\usepackage{tikz}
\usetikzlibrary{arrows.meta,positioning,fit,shapes.misc}
\usepackage{comment}
\usepackage[none]{hyphenat} 
\usepackage{microtype}      
\def\BibTeX{{\rm B\kern-.05em{\sc i\kern-.025em b}\kern-.08em
		T\kern-.1667em\lower.7ex\hbox{E}\kern-.125emX}}
\begin{document}
	
\title{Dynamic Multi-Depot Vehicle Routing with Online Requests:
Event-Driven Transformer--DRL and Rolling-Horizon Benchmarking}

\author{Faezeh Ardali and Gerald M. Knapp%
\thanks{Faezeh Ardali and Gerald M. Knapp are with the Department of
Industrial Engineering, Louisiana State University, Baton Rouge, LA, USA
(e-mail: fardal1@lsu.edu; gknapp@lsu.edu).}
}

	\maketitle
\begin{abstract}
This paper presents an event-driven learning and benchmarking framework for the Dynamic Multi-Depot Vehicle Routing Problem with progressively revealed requests and evolving vehicle states. Masked MLP and Transformer policies are trained through behavior cloning and proximal policy optimization. Deterministic feasibility masking prevents invalid vehicle--request assignments, while fixed-prefix/flexible-suffix route commitments protect completed, active, and near-term decisions and separately measure vehicle reassignment and resequencing. The learned policies are compared with dynamic insertion heuristics and time-limited rolling-horizon optimization. In a 20-scenario policy benchmark, all methods completed every request without invalid actions, but nearest feasible achieved the lowest mean objective and outperformed the learned policies in routing quality, waiting time, stability, makespan, and runtime. Across five independent training runs, PPO had little average effect on the MLP and improved the Transformer on average, although with greater seed variability. Under the common protocol, nearest feasible achieved the lowest combined objective and route disruption, whereas rolling horizon achieved the lowest waiting times and makespan at substantially higher computational cost. The learned policies retained millisecond-level decisions and transferred to instances with up to 80 requests without retraining, but did not outperform the strongest heuristic. No single method was best across routing efficiency, service responsiveness, stability, and online computation.
\end{abstract}


\section{Introduction}

\IEEEPARstart{V}{ehicle} routing is a fundamental decision problem in transportation, distribution, and service operations. It determines how a fleet of vehicles should serve geographically distributed customers while satisfying requirements such as vehicle capacity, service demand, route duration, and depot assignment. Vehicle Routing Problems (VRPs) arise in applications including last-mile delivery, field service, technician dispatch, mobile healthcare, and emergency logistics~\cite{konstantakopoulos2022vehicle}. Classical VRPs assume that all requests are known before dispatch, whereas practical systems often receive new requests after vehicle operations have begun~\cite{zhang2025pickup}.
The Dynamic Vehicle Routing Problem (DVRP) addresses this setting by allowing requests and operational states to evolve over time~\cite{pan2023deep}.
In the multi-depot case, the dispatcher must repeatedly determine which vehicle and depot should serve each request while accounting for changing vehicle locations, capacities, service statuses, and customer waiting times. The resulting D-MDVRP is therefore a sequential decision problem rather than a single static routing task.

\subsection{Related Work and Positioning}

Dynamic routing methods commonly respond to newly revealed requests through local insertion, event-driven or periodic replanning, and rolling-horizon reoptimization~\cite{el2025metaheuristic,meraliyev2025comprehensive,zhang2025pickup}. Beyond classical approaches, emerging quantum optimization frameworks have also targeted combinatorial engineering problems through QUBO-based formulations~\cite{rajabi2026distributed}; while promising, their practical use for many real-world engineering problems remains constrained. Fast insertion rules revise routes locally, whereas rolling-horizon methods repeatedly reoptimize the currently available requests and constraints at successive decision epochs, generally at greater online computational cost~\cite{zhang2025pickup}.

Learning-based dispatch provides an alternative in which policies are trained offline and evaluated rapidly during operation. Such approaches have been applied to dynamic and uncertain VRPs, post-disaster damage assessment and repair scheduling, event-driven crew dispatch, and autonomous resource-allocation problems~\cite{amani2026event,torkaman2026integrated,soleymani2024simulation}. Related work has also applied Transformer-based learning to combinatorial scheduling and evaluated transfer to larger problem instances without retraining~\cite{ardali2026deep}. Transformer architectures use attention to model dependencies among encoded elements, and related dispatch work has used them to capture high-dimensional system states and temporal dependencies~\cite{vaswani2017attention,amani2026learning}. In the present setting, however, online deployment also requires explicit feasibility handling and repeated route reconstruction as new information arrives.

Disruption-management studies further emphasize that real-time replanning should balance operational quality against deviations from an existing plan~\cite{eglese2018disruption,gmira2021managing}. Related rolling-horizon scheduling work has similarly combined near-term freeze windows with resequencing penalties to balance responsiveness and schedule stability under evolving information~\cite{parmar2026smart}. Motivated by these considerations, the present framework protects completed, active, and near-term decisions and reports vehicle reassignment and predecessor change separately rather than treating every route modification as an undifferentiated change.

The contribution is an integrated framework rather than a new routing, attention, or reinforcement-learning algorithm. It combines an event-driven D-MDVRP environment, deterministic feasibility masking, behavior-cloned MLP and Transformer policies with PPO fine-tuning, fixed-prefix and flexible-suffix route commitments, and separate reassignment and resequencing measures. Heuristics, learned policies, and rolling-horizon optimization are then evaluated under the same feasibility, commitment, objective, scenario, and runtime protocol. Its main distinction is the common stability-aware comparison of components usually studied separately.

The main contributions of this paper are an event-driven D-MDVRP environment with deterministic feasibility masking, MLP and Transformer policies trained using behavior cloning and PPO, fixed-prefix/flexible-suffix commitments with separate reassignment and resequencing measures, and common-protocol comparisons with heuristics and rolling-horizon optimization.



\section{Dynamic Routing Model and Assumptions}
\label{sec:dynamic_context}

We consider a dynamic service-routing environment in which capacitated vehicles operate from multiple depots. An initial subset of requests is known before dispatch, while additional requests are revealed during route execution. The routing state is updated after request arrivals, vehicle arrivals, service starts, and service completions. Completed services, active movements, and near-term committed decisions remain fixed, while eligible requests in the flexible route portions may be reassigned or reordered.

\subsection{Dynamic Multi-Depot Service Network}
\label{subsec:network}

Let \(\mathcal{D}\), \(\mathcal{K}\), and \(\mathcal{N}\) denote the depot, vehicle, and request sets, respectively. Vehicle \(k\in\mathcal{K}\) has home depot \(d(k)\in\mathcal{D}\) and capacity \(Q_k\), while \(\mathcal{N}_t\subseteq\mathcal{N}\) contains the requests revealed by time \(t\).

Each request \(i\in\mathcal{N}\) is represented by the 10-dimensional normalized feature vector
\begin{equation}
\small
\begin{aligned}
\widetilde{\xi}_i(t)
=\bigg[
&\frac{x_i}{L},\,
\frac{y_i}{L},\,
\frac{q_i}{Q_{\max}},\,
\frac{s_i}{15},\,
\frac{p_i}{3},\,
\frac{a_i}{T_{\mathrm{hor}}},\\
&\frac{\min\{\max(t-a_i,0),T_{\mathrm{hor}}\}}
{T_{\mathrm{hor}}},\,
\frac{c_i^{\mathrm{stat}}(t)}{5},\,
\bar{k}_i(t),\,
\delta_i^{\mathrm{pen}}(t)
\bigg].
\end{aligned}
\label{eq:request_features}
\end{equation}
Here, \(L\), \(Q_{\max}\), and \(T_{\mathrm{hor}}\) are the coordinate, capacity, and time scales. The features describe location, demand, service duration, priority, arrival time, waiting time, request status, assigned vehicle, and pending status. The status code \(c_i^{\mathrm{stat}}(t)\in\{0,\ldots,5\}\) represents unrevealed, pending, assigned, traveling, serving, and completed requests. The assigned-vehicle feature is zero when unassigned and otherwise equals
\((k_i^{\mathrm{asg}}(t)+1)/|\mathcal{K}|\), while
\(\delta_i^{\mathrm{pen}}(t)\) equals one only for pending requests. Unrevealed requests remain in the fixed-size representation but are excluded by the padding and feasibility masks.

Vehicle \(k\) is represented by the nine-dimensional feature vector
\begin{equation}
\small
\begin{aligned}
\widetilde{\nu}_k(t)
=\bigg[
&\frac{x_k(t)}{L},\,
\frac{y_k(t)}{L},\,
\frac{Q_k^{\mathrm{rem}}(t)}{Q_k},\,
\frac{Q_k^{\mathrm{used}}(t)}{Q_k},\,
\frac{c_k^{\mathrm{stat}}(t)}{2},\\
&\bar{r}_k^{\mathrm{cur}}(t),\,
\frac{n_k(t)}{N},\,
\frac{d(k)}{\max\{1,|\mathcal{D}|-1\}},\,
\frac{D_k(t)}{LN}
\bigg].
\end{aligned}
\label{eq:vehicle_features}
\end{equation}
The features describe current location, remaining and used capacity, operating status, active request, flexible-route length, home depot, and cumulative distance. The vehicle-status code \(c_k^{\mathrm{stat}}(t)\in\{0,1,2\}\) represents idle, traveling, and serving states. The current-request feature is zero when no request is active and otherwise equals
\((r_k^{\mathrm{cur}}(t)+1)/N\).

Each vehicle route begins and ends at its home depot and contains its ordered assigned requests. Executed, active, and committed portions remain fixed, whereas the flexible suffix may be revised during replanning. Euclidean distance and the corresponding constant-speed travel time are used in the numerical experiments. Request attributes and arrival times follow the scenario settings described in Section~\ref{sec:case_study}.

\subsection{Event-Driven Requests and Route Commitments}
\label{subsec:event_commitment}

New requests are added to the revealed set at their arrival events. Each revealed request is pending, assigned, active, or completed. Assigned requests are further classified as fixed or flexible. The fixed set contains the next protected request of each vehicle and any request scheduled to begin service within the commitment horizon \(H^{\mathrm{commit}}\); active travel and service are protected automatically.

Each route is therefore represented by a fixed prefix and a flexible suffix,
\(\mathcal{P}_k(t)=\mathcal{P}_k^{\mathrm{fix}}(t)\oplus
\mathcal{P}_k^{\mathrm{flex}}(t)\).
Only the flexible suffix may be reassigned or reordered. At each operational event, the environment updates request statuses, vehicle locations, capacities, availability, waiting times, and remaining routes. Common scenario seeds ensure identical operating conditions across methods.

\section{Event-Driven Learning Framework for Dynamic Vehicle Routing}
\label{sec:methodology}

The D-MDVRP is formulated as an event-driven sequential decision problem. Under common state, action, and reward definitions, deterministic feasibility and route-commitment rules are combined with MLP and Transformer policies trained by behavior cloning and PPO.

\subsection{Sequential Decision Model}
\label{subsec:mdp}

The routing environment is a finite-horizon Markov decision process whose state contains the request, vehicle, depot, and global features. The global features are normalized simulation time, revealed, completed, and pending-request proportions, replanning count, and cumulative route changes.

Requests progress through the unrevealed, pending, assigned, traveling, serving, and completed states. At a replan, fixed prefixes remain unchanged and requests in the previous flexible suffix return to the pending set. Each action assigns one pending request and appends it to a vehicle's reconstructed suffix. Construction ends when no revealed pending request remains; unrevealed requests wait until arrival, and there is no separate stop or defer action.

Vehicle load is initialized from completed, active, and committed demand and updated after every assignment. The mask prevents capacity violations and preserves sufficient residual fleet capacity. If pending requests remain but all actions are masked, the state is declared infeasible. Previous vehicle assignments and predecessors are stored before replanning; newly revealed requests incur no change penalty, whereas changes to previously planned requests are finalized after reconstruction.

\subsection{Masked MLP and Transformer Policies}
\label{subsec:policy}

At each decision, the actor scores every vehicle--request pair. The base instances contain four vehicles and 30 requests, producing 120 candidate actions. Invalid logits are set to negative infinity before softmax or greedy selection. A pair is valid only if the request is revealed, pending or flexibly assigned, not completed, active, committed, or previously selected during the current reconstruction, and feasible for the selected vehicle. Invalid actions therefore receive zero probability during BC, PPO, validation, and testing.

\subsubsection{MLP Actor and Critic}

The MLP actor represents each pair using 10 request, nine vehicle, six global, and five pair-specific features. The pair-specific features are nearest distance, incremental distance, predicted waiting time, route-change indicator, and route length. A shared \(30\!\rightarrow\!64\!\rightarrow\!32\!\rightarrow\!1\) ReLU network scores all pairs. The critic pools the elementwise mean and maximum of feasible candidate vectors and applies a \(60\!\rightarrow\!64\!\rightarrow\!32\!\rightarrow\!1\) Tanh network.

\subsubsection{Transformer Actor and Critic}

The Transformer uses 30 request, four vehicle, and two depot tokens with feature dimensions 10, nine, and six, respectively. Depot \(d\) is represented by
\begin{equation}
\small
\left[
\frac{x_d}{L},\,
\frac{y_d}{L},\,
\frac{d}{\max\{1,|\mathcal{D}|-1\}},\,
\frac{n_d^{\mathrm{home}}}{|\mathcal{K}|},\,
\frac{t}{T_{\mathrm{hor}}},\,
1
\right],
\end{equation}
where \(n_d^{\mathrm{home}}\) is the number of vehicles based at depot \(d\). Separate linear projections map the request, vehicle, and depot features to \(d_{\mathrm{model}}=32\). Learned token-type and position embeddings and a projected six-dimensional global vector are added to the tokens. The encoder has one layer, four attention heads, feedforward dimension 64, ReLU activation, and zero dropout; unrevealed requests are zeroed and excluded through the padding mask.

For each vehicle--request pair, the encoded vehicle, request, and global vectors and the five edge features form a 101-dimensional vector, which is processed by a shared \(101\!\rightarrow\!64\!\rightarrow\!1\) ReLU scorer. The Transformer critic uses the elementwise mean and maximum of the visible 10-dimensional request features, all nine-dimensional vehicle features, all six-dimensional depot features, and the five-dimensional features of feasible edges, together with the six global features. The resulting dimension is \(2(10+9+6+5)+6=66,\) and the critic applies a \(66\!\rightarrow\!64\!\rightarrow\!32\!\rightarrow\!1\) Tanh network.

\subsection{Route Commitment, Stability, Objective, and Reward}
\label{subsec:stability_reward}

For the principal heuristic and neural-policy benchmark, the evaluation objective is
\begin{equation}
\small
J_{\mathrm{eval}}
=
D+0.10W+2.00N_{\mathrm{chg}},
\label{eq:combined_evaluation_objective}
\end{equation}
where \(D\) is total executed distance including depot returns, \(W\) is total customer waiting time in customer-minutes, and \(N_{\mathrm{chg}}\) counts previously planned flexible requests whose vehicle--predecessor pair changes at a replanning event. The coefficients have units of distance per customer-minute and distance per route change. Average and maximum waiting, makespan, completion, and runtime are reported separately.

PPO uses incremental distance, predicted waiting, and the corresponding action-level change indicator:
{\small
\begin{align}
r_t^{\mathrm{dense}}
&=
-0.01\left(
\Delta D_t+0.10\widehat{w}_{i,t}+2.00I_t^{\mathrm{chg}}
\right),
\nonumber\\
r_T^{\mathrm{corr}}
&=
-0.01J_{\mathrm{eval}}
-\sum_{t=0}^{T}r_t^{\mathrm{dense}},
\qquad
\sum_{t=0}^{T}r_t=-0.01J_{\mathrm{eval}}.
\label{eq:reward_objective_alignment}
\end{align}}
Thus, dense rewards provide intermediate feedback while the undiscounted episode reward remains exactly aligned with the evaluation objective. PPO uses \(\gamma=0.99\), so the timing of intermediate costs may affect the discounted return.

For the commitment-aware comparison, disruption is separated into vehicle reassignment and predecessor change:
\begin{equation}
J_{\mathrm{stab}}
=
D+0.10W+2.00N_{\mathrm{asg}}+1.00N_{\mathrm{seq}},
\label{eq:commitment_combined_objective}
\end{equation}
where \(N_{\mathrm{asg}}\) and \(N_{\mathrm{seq}}\) count flexible requests whose vehicle and predecessor change, respectively. A request may contribute to both terms.


\subsection{Behavior Cloning and PPO Fine-Tuning}
\label{subsec:training}

The MLP and Transformer policies are trained by behavior cloning (BC) followed by PPO fine-tuning.

\subsubsection{Behavior-Cloning Dataset Construction}

Each BC sample contains the complete state, deterministic mask, and expert vehicle--request action; one sample is recorded per expert decision, and all data splits are scenario-disjoint. The MLP expert is waiting-aware insertion. Its training, validation, and held-out seeds are \(1000\)--\(1011\), \(2000\)--\(2003\), and \(3000\)--\(3005\), yielding 4,525 training and 1,528 validation samples.

At each decision, the Transformer expert first forms the set \(\mathcal{C}_t\) of unique actions proposed by nearest feasible, cheapest append, and waiting-aware insertion. It then selects the candidate \((k,i)\in\mathcal{C}_t\) minimizing
{\small
\begin{align}
S_{k,i}^{\mathrm{hyb}}
={}&
0.55\frac{\Delta d_{k,i}}{L}
+
0.30(0.75+0.25p_i)
\frac{\widehat{w}_{k,i}}{T_{\mathrm{hor}}}
\nonumber\\
&+
0.10I_{k,i}^{\mathrm{chg}}
+
0.04\frac{Q_k^{\mathrm{plan}}}{Q_k}
+
0.01\frac{n_k}{N},
\label{eq:hybrid_expert_score}
\end{align}}
where \(\Delta d_{k,i}\) is incremental distance, \(\widehat{w}_{k,i}\) is predicted waiting time, \(I_{k,i}^{\mathrm{chg}}\) is the route-change indicator, \(Q_k^{\mathrm{plan}}\) is the current planned cumulative load, and \(n_k\) is the planned route length. Scores are rounded to 12 decimal places for deterministic comparison, and ties are resolved using the smallest flattened vehicle--request action index. Transformer training, validation, and held-out seeds are \(4100\)--\(4111\), \(5100\)--\(5103\), and \(6100\)--\(6105\), yielding 4,266 training and 1,372 validation samples.

The actors use masked cross-entropy and retain the checkpoint with minimum validation cross-entropy. Table~\ref{tab:bc_training_settings} summarizes the principal BC settings.

\begin{table}[t]
\centering
\caption{Behavior-cloning configurations.}
\vspace{-3pt}
\label{tab:bc_training_settings}
\vspace{-6pt}
\scriptsize
\renewcommand{\arraystretch}{1.06}
\setlength{\tabcolsep}{4.0pt}
\begin{tabular}{lcc}
\toprule
\textbf{Setting} & \textbf{MLP} & \textbf{Transformer} \\
\midrule
Expert
& Waiting-aware
& Hybrid \\

Scenarios (train/val./test)
& \(12/4/6\)
& \(12/4/6\) \\

Samples (train/val.)
& \(4525/1528\)
& \(4266/1372\) \\

Epochs/batch size
& \(12/128\)
& \(10/128\) \\

Adam learning rate
& \(2.0\times10^{-3}\)
& \(1.5\times10^{-3}\) \\

Maximum gradient norm
& \(5.0\)
& \(5.0\) \\
\bottomrule
\end{tabular}
\vspace{-2pt}
\end{table}

\subsubsection{PPO Fine-Tuning}

The selected BC actor initializes PPO. Separate Adam optimizers are used for the actor and critic, with generalized advantage estimation, normalized advantages, and the standard clipped PPO loss. Table~\ref{tab:ppo_training_settings} reports the complete principal settings.

\begin{table}[t]
\centering
\caption{PPO training configurations.}
\vspace{-5pt}
\label{tab:ppo_training_settings}
\vspace{-2pt}
\scriptsize
\renewcommand{\arraystretch}{1.05}
\setlength{\tabcolsep}{3.5pt}
\begin{tabular}{lcc}
\toprule
\textbf{Setting} & \textbf{MLP--PPO} & \textbf{Trans.--PPO} \\
\midrule
Iterations/episodes/epochs
& \(8/3/3\)
& \(6/2/2\) \\

Episodes (train/val./test)
& \(24/4/6\)
& \(12/4/6\) \\

Minibatch size
& \(256\)
& \(128\) \\

Actor/critic learning rate
& \(2{\times}10^{-4}/1{\times}10^{-3}\)
& \(2{\times}10^{-4}/8{\times}10^{-4}\) \\

\(\gamma/\lambda/\epsilon\)
& \(0.99/0.95/0.15\)
& \(0.99/0.95/0.12\) \\

Value/entropy coefficient
& \(0.50/0.005\)
& \(0.50/0.001\) \\

Gradient norm/reward scale
& \(1.0/0.01\)
& \(1.0/0.01\) \\

Critic hidden layers
& \(64,32\)
& \(64,32\) \\
\bottomrule
\end{tabular}
\vspace{-2pt}
\end{table}

MLP training, validation, and held-out seeds are \(4000\)--\(4023\), \(5000\)--\(5003\), and \(6000\)--\(6005\); the corresponding Transformer seeds are \(7200\)--\(7211\), \(8200\)--\(8203\), and \(9200\)--\(9205\). After each iteration, greedy validation retains the actor--critic pair with the lowest mean objective. BC is iteration zero, so PPO replaces it only after validation improvement; held-out and common-protocol scenarios are excluded from checkpoint selection.
\subsection{Event-Driven Operation and Rolling-Horizon Benchmark}
\label{subsec:event_driven}

The simulator advances directly among request arrivals, vehicle arrivals, service starts, and service completions. When replanning is required, active and committed prefixes are retained, flexible requests return to the pending set, and cumulative loads are initialized from completed, active, and committed demand. Feasible actions are then selected until no revealed pending request remains. The reconstructed suffixes and their assignment and sequence changes are finalized before operations resume.

Training is offline. During testing, a learned policy requires only the current state, one forward evaluation per decision, and deterministic feasibility checks; request arrivals do not trigger retraining.

The rolling- baseline uses the same event states and fixes completed, active, and committed route segments. Flexible requests may be reassigned and resequenced subject to capacity constraints. Each solve is limited to one second, and the best feasible incumbent returned within this limit is implemented. These incumbents do not necessarily represent proven optimal solutions. Solver optimality-gap statistics were not recorded and are therefore unavailable.

\section{Numerical Results}
\label{sec:case_study}

This section evaluates the event-driven dynamic routing framework in terms of solution feasibility, routing performance, scalability, route stability, and online computation time. The experiments compare dynamic insertion heuristics, behavior-cloned neural policies, PPO-fine-tuned policies, and a time-limited rolling-horizon optimization baseline.

\subsection{Experimental Setup}
\label{subsec:experimental_setup}

Synthetic Euclidean instances use vehicle capacity 30 and fixed scenario seeds. Small, medium, and large cases contain \((2,4,18,12)\), \((3,6,30,20)\), and \((4,8,48,32)\) depots, vehicles, initial requests, and dynamic requests, respectively, giving 30, 50, and 80 total requests. Dynamic requests arrive in four waves. Scaling results are averaged over three scenarios per size, using seeds \(12000\)--\(12002\), \(13000\)--\(13002\), and \(14000\)--\(14002\) for the small, medium, and large cases, respectively. The MLP is applied directly across sizes. For the Transformer, the request, vehicle, and depot blocks of the learned position embeddings are linearly interpolated separately, while all projection, attention, and scoring weights remain unchanged; no scale-specific retraining is performed.

The unified heuristic and neural-policy benchmark uses 20 unseen scenarios with seeds \(10000\)--\(10019\) and the objective in \eqref{eq:combined_evaluation_objective}. The common-protocol experiment reuses the same scenario seeds and generated request realizations but evaluates nearest feasible, waiting-aware, MLP--PPO, Transformer--PPO, and rolling horizon in the commitment-aware environment with a 15-minute commitment horizon and the objective in \eqref{eq:commitment_combined_objective}. Consequently, differences between the two result tables arise from the commitment and stability protocol rather than from different scenario samples. Previously selected neural checkpoints are applied without retraining.

Experiments use Python~3.8.4 and PyTorch~2.4.1 on an 11th Gen Intel(R) Core(TM) i7-1165G7 CPU at 2.80~GHz with 32~GB RAM, without GPU acceleration. Rolling horizon uses OR-Tools~9.8.3296 with SCIP~8.0.4, one thread, and a one-second limit. End-to-end replanning time includes state construction, all sequential decisions, masking, route updates, and stability accounting; rolling-horizon timing additionally includes model construction, optimization, and route extraction. Training, model loading, and simulated operations are excluded. The timing set contains 3,726 replanning events, and every rolling-horizon call returned a feasible incumbent. Source code, experiment scripts, and reference outputs used in this study are publicly available in the accompanying GitHub repository~\cite{ardali2026dmdvrpcode}.

Random selects uniformly from feasible pairs; nearest feasible minimizes endpoint-to-request distance; cheapest append minimizes incremental distance; and waiting-aware minimizes incremental distance plus \(0.10\) times predicted waiting and \(2.00\) times the change indicator. The Hybrid expert uses the weighted score defined in the training subsection. Waiting-aware and Hybrid were selected as BC teachers before final testing to provide multi-criteria demonstrations rather than to reproduce the retrospectively strongest test heuristic.

\subsection{Unified Comparison of Heuristic and Neural Policies}
\label{subsec:overall_results}

Table~\ref{tab:unified_benchmark} reports the mean and sample standard deviation over 20 common test scenarios. All methods achieved 100\% request completion under the common feasibility rules, and no invalid actions were selected.

\begin{table}[!t]
\centering
\caption{Unified routing benchmark.}
\vspace{-5pt}
\label{tab:unified_benchmark}
\vspace{-3pt}

\fontsize{6.5}{7.3}\selectfont
\renewcommand{\arraystretch}{1.05}

\begin{tabular*}{\columnwidth}{@{\extracolsep{\fill}}lcccc@{}}
\toprule
\multicolumn{5}{c}{\textit{(a) Routing and service performance}} \\
\midrule
\textbf{Method} &
\textbf{Distance} &
\textbf{Avg. wait} &
\textbf{Max wait} &
\textbf{Objective} \\
\midrule

Random
& $384.0 \pm 30.4$
& $49.4 \pm 6.9$
& $135.5 \pm 22.1$
& $1036.1 \pm 147.5$ \\

Nearest
& $179.1 \pm 14.1$
& $44.5 \pm 7.0$
& $134.9 \pm 20.9$
& $346.1 \pm 39.5$ \\

Cheapest
& $191.1 \pm 20.9$
& $62.6 \pm 10.9$
& $196.1 \pm 33.3$
& $477.5 \pm 72.1$ \\

Waiting-aware
& $193.4 \pm 22.8$
& $55.6 \pm 9.0$
& $178.3 \pm 30.0$
& $428.7 \pm 52.5$ \\

Hybrid
& $190.8 \pm 13.7$
& $58.3 \pm 7.8$
& $193.8 \pm 28.6$
& $434.9 \pm 41.0$ \\

MLP--BC
& $194.4 \pm 19.3$
& $56.3 \pm 10.0$
& $183.4 \pm 26.7$
& $432.9 \pm 58.8$ \\

MLP--PPO
& $192.0 \pm 18.6$
& $56.8 \pm 10.1$
& $188.5 \pm 27.3$
& $430.8 \pm 57.5$ \\

Trans.--BC
& $195.1 \pm 19.2$
& $60.3 \pm 7.4$
& $193.2 \pm 25.9$
& $431.2 \pm 42.5$ \\

Trans.--PPO
& $200.9 \pm 17.8$
& $60.2 \pm 7.0$
& $192.1 \pm 20.5$
& $441.9 \pm 37.7$ \\

\bottomrule
\end{tabular*}

\vspace{4pt}

\begin{tabular*}{\columnwidth}{@{\extracolsep{\fill}}lccc@{}}
\toprule
\multicolumn{4}{c}{\textit{(b) Operational stability and runtime}} \\
\midrule
\textbf{Method} &
\textbf{Makespan} &
\textbf{Changes} &
\textbf{Time} \\
&
\textbf{(min)} &
&
\textbf{(ms/action)} \\
\midrule

Random
& $188.5 \pm 12.6$
& $252.0 \pm 58.4$
& $0.092 \pm 0.012$ \\

Nearest
& $187.7 \pm 23.7$
& $16.9 \pm 12.2$
& $0.156 \pm 0.022$ \\

Cheapest
& $221.4 \pm 32.7$
& $49.3 \pm 25.6$
& $0.163 \pm 0.022$ \\

Waiting-aware
& $202.8 \pm 19.5$
& $34.3 \pm 10.2$
& $0.164 \pm 0.023$ \\

Hybrid
& $210.8 \pm 25.9$
& $34.6 \pm 10.3$
& $0.495 \pm 0.068$ \\

MLP--BC
& $202.1 \pm 18.2$
& $34.8 \pm 12.6$
& $0.538 \pm 0.029$ \\

MLP--PPO
& $206.0 \pm 19.8$
& $34.2 \pm 12.3$
& $0.609 \pm 0.026$ \\

Trans.--BC
& $208.4 \pm 23.3$
& $27.6 \pm 9.3$
& $0.867 \pm 0.045$ \\

Trans.--PPO
& $204.4 \pm 18.6$
& $30.2 \pm 7.5$
& $0.763 \pm 0.028$ \\

\bottomrule
\end{tabular*}

\vspace{1pt}

\parbox{\columnwidth}{
\fontsize{6.5}{7.3}\selectfont
BC denotes behavior cloning; Trans. denotes Transformer.
}
\end{table}

Nearest feasible achieved the lowest objective, \(346.13\), was best in 18 of 20 scenarios, and outperformed all learned policies in distance, waiting, makespan, route changes, and runtime. Its mean runtime was \(0.156\)~ms/action versus \(0.538\)--\(0.867\)~ms/action for the learned policies. In these tests, the learned policies offered no quality or runtime advantage over the nearest feasible.

For the selected checkpoints, MLP--PPO improved the BC objective by \(0.47\%\), whereas Transformer--PPO worsened it by \(2.50\%\). Across the five independent runs discussed next, the corresponding average PPO changes were \(0.13\%\) for the MLP and a \(2.50\%\) improvement for the Transformer, with higher Transformer variability. The learned policies instead offer feasible millisecond inference and limited size transfer, not better performance than the strongest heuristic.

\subsection{Robustness Across Independent Training Seeds}
\label{subsec:training_seed_robustness}

The complete BC--PPO pipeline was repeated for seeds \(2031\)--\(2035\) using fixed training, validation, and test scenarios, thereby isolating initialization, minibatch, and action-sampling variability. Each selected policy was evaluated on the same 20 unseen scenarios.

\begin{figure}[t]
\centering
\includegraphics[width=0.8\columnwidth]{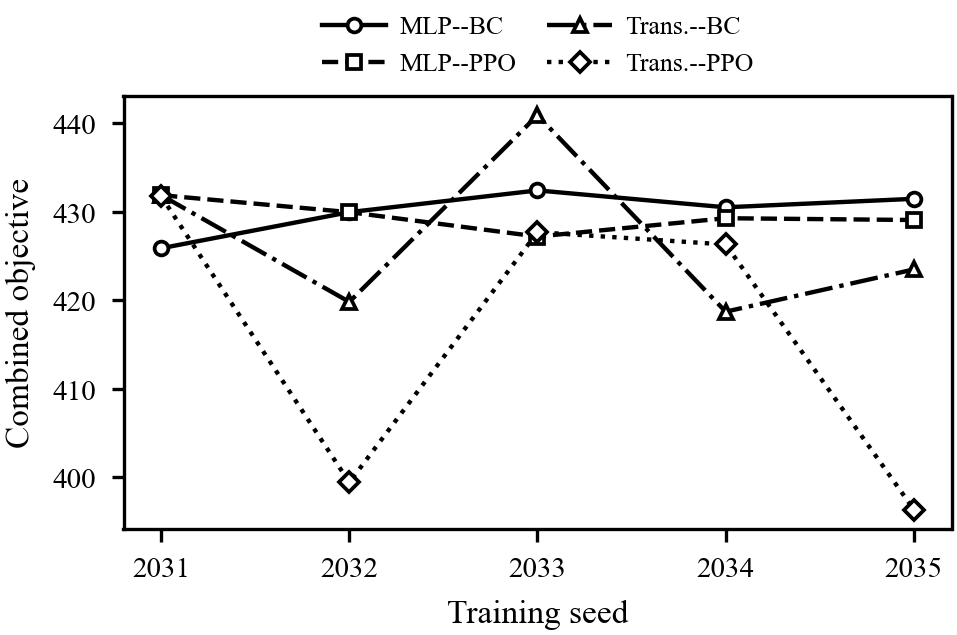}
\vspace{-5pt}
\caption{Combined objective across five independent BC--PPO training runs.}
\label{fig:training_seed_robustness}
\vspace{-6pt}
\end{figure}

All 400 evaluations completed every request without invalid actions. Mean objectives were \(430.057\pm2.513\) for MLP--BC, \(429.489\pm1.689\) for MLP--PPO, \(426.968\pm9.358\) for Transformer--BC, and \(416.308\pm16.983\) for Transformer--PPO. PPO changed the MLP little on average and improved the Transformer by \(2.50\%\), with greater seed variability. Nearest feasible remained best.

\subsection{Ablation and Statistical Analysis}
\label{subsec:ablation_analysis}

The waiting-aware policy was evaluated on the same 20 scenarios using the full framework and variants without near-term commitment, reassignment penalty, or resequencing penalty. All variants were evaluated using the common reference objective \(D+0.10W+2.00N_{\mathrm{asg}}+1.00N_{\mathrm{seq}}\), regardless of the decision-rule weights. 

\begin{table}[t]
\centering
\caption{Ablation of route commitment and stability penalties.}
\vspace{-5pt}
\label{tab:stability_ablation}
\vspace{-1pt}

\fontsize{6.2}{6.8}\selectfont
\renewcommand{\arraystretch}{1.04}
\setlength{\tabcolsep}{2.0pt}

\begin{tabularx}{\columnwidth}{
@{}
>{\raggedright\arraybackslash}X
cccc
@{}
}
\toprule
\textbf{Configuration} &
\textbf{Objective} &
\shortstack{\textbf{Reassign.}} &
\shortstack{\textbf{Sequence}} &
\shortstack{\textbf{Adj.}\\\(\boldsymbol{p}\)\textbf{-value}} \\
\midrule

Full framework
& \(414.38\pm50.06\)
& \(14.50\pm9.29\)
& \(24.30\pm8.29\)
& -- \\

No near-term commit.
& \(434.37\pm58.50\)
& \(18.50\pm9.08\)
& \(29.60\pm10.41\)
& 0.824 \\

No reassignment penalty
& \(430.29\pm53.74\)
& \(19.95\pm11.46\)
& \(26.95\pm10.22\)
& \(\mathbf{0.007}\) \\

No resequencing penalty
& \(427.84\pm51.93\)
& \(16.35\pm8.38\)
& \(27.95\pm9.96\)
& 0.098 \\

\bottomrule
\end{tabularx}

\vspace{1pt}
\parbox{\columnwidth}{
\fontsize{6.2}{6.8}\selectfont
Values are mean \(\pm\) SD over 20 paired scenarios. Adjusted \(p\)-values use two-sided exact sign tests against the full framework with Holm correction.
}
\vspace{-3pt}
\end{table}

The full framework achieved the lowest mean objective. Removing commitment, reassignment penalty, and resequencing penalty increased it by \(4.83\%\), \(3.84\%\), and \(3.25\%\), respectively; only removal of the reassignment penalty remained significant after Holm correction (\(p=0.007\)).

Exact Wilcoxon tests across the five training runs found no significant differences for MLP--BC versus MLP--PPO (\(p=0.875\)), Transformer--BC versus Transformer--PPO (\(p=0.188\)), or MLP--PPO versus Transformer--PPO (\(p=0.188\)). These five-pair tests have limited power, so PPO and architecture effects remain seed-dependent.
\subsection{Scenario-Scaling Performance}
\label{subsec:scalability}

\begin{figure}[!t]
\centering
\includegraphics[width=0.75\columnwidth]
{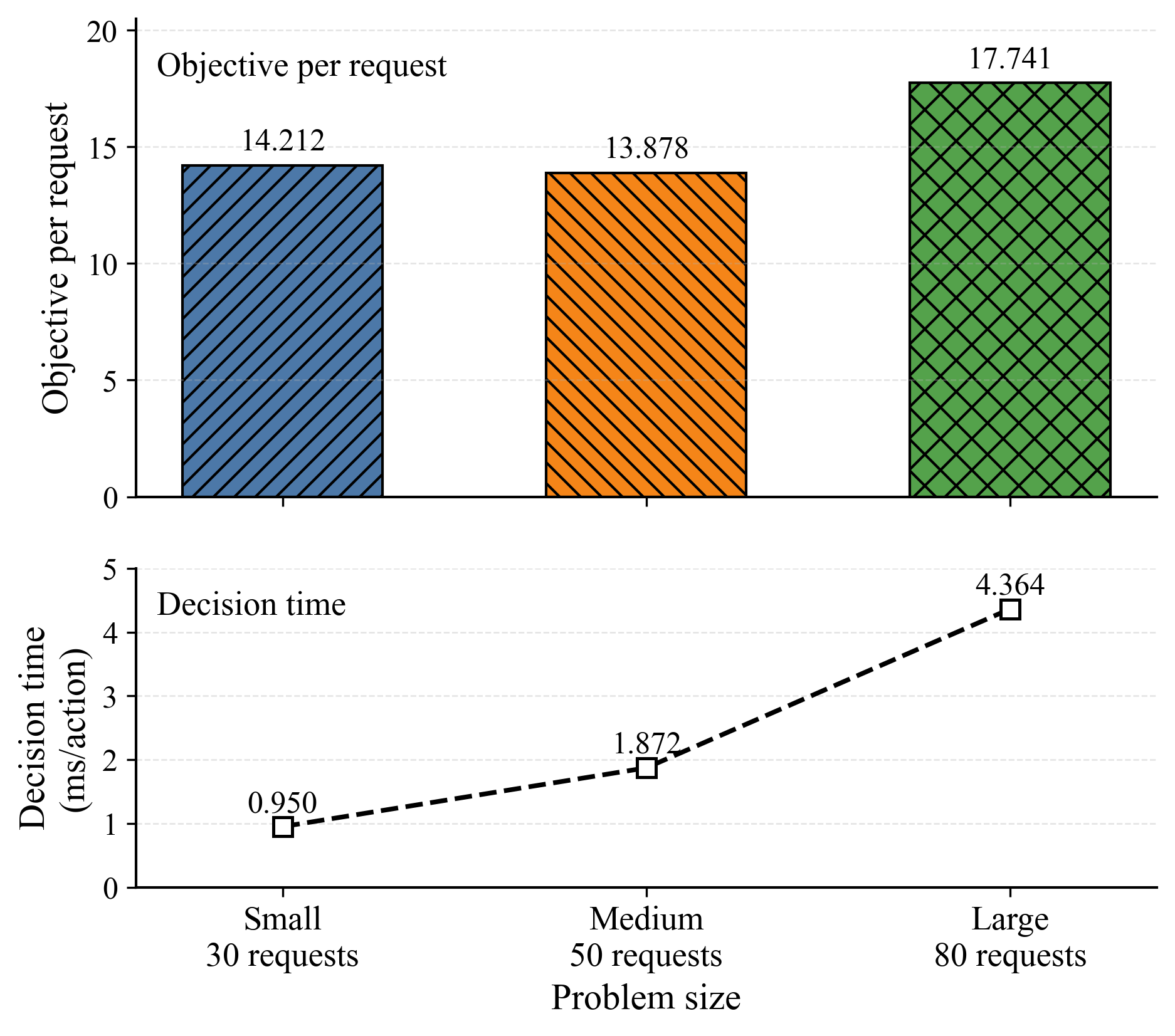}
\vspace{-5pt}
\caption{Transformer--PPO objective and decision time across problem sizes.}
\label{fig:transformer_scaling}
\vspace{-4pt}
\end{figure}

The trained policies were applied without retraining to instances containing 30, 50, and 80 requests. All methods completed every request and satisfied the implemented capacity constraints. Table~\ref{tab:scaling_results} provides the comparison with the strongest heuristic, while Fig.~\ref{fig:transformer_scaling} highlights Transformer--PPO size transfer.

\begin{table}[!t]
\centering
\caption{Scaling means over three scenarios per size: objective per request/runtime in ms per action.}
\vspace{-4pt}
\label{tab:scaling_results}
\vspace{-7pt}
\scriptsize
\renewcommand{\arraystretch}{1.07}
\setlength{\tabcolsep}{5.0pt}
\begin{tabular}{lccc}
\toprule
\textbf{Method} &
\textbf{Small} &
\textbf{Medium} &
\textbf{Large} \\
\midrule
Nearest
& \(10.70/0.21\)
& \(12.08/0.61\)
& \(14.35/1.72\) \\

MLP--PPO
& \(13.70/0.67\)
& \(13.25/1.49\)
& \(16.25/3.54\) \\

Transformer--PPO
& \(14.21/0.95\)
& \(13.88/1.87\)
& \(17.74/4.36\) \\
\bottomrule
\end{tabular}
\vspace{-2pt}
\end{table}

Nearest feasible retained the lowest objective and runtime at every scale. Transformer--PPO completed all instances without retraining, but its objective per request increased from \(14.21\) to \(17.74\) and its runtime from \(0.95\) to \(4.36\)~ms/action as the instance size increased from 30 to 80 requests. This shows feasible size transfer over the tested range, but no advantage over nearest feasible.

\begin{table*}[!t]
\centering
\caption{Common-protocol results over 20 paired scenarios (mean \(\pm\) sample SD).}
\vspace{-5pt}
\label{tab:common_protocol_results}
\vspace{-6pt}
\fontsize{6.2}{6.9}\selectfont
\setlength{\tabcolsep}{2.8pt}
\renewcommand{\arraystretch}{1.08}
\begin{tabular*}{\textwidth}{@{\extracolsep{\fill}}lcccccc@{}}
\toprule
\textbf{Method} &
\textbf{Distance} &
\shortstack{\textbf{Avg./max wait}\\\textbf{(min)}} &
\textbf{Makespan} &
\textbf{Objective} &
\shortstack{\textbf{Reassign./}\\\textbf{sequence}} &
\shortstack{\textbf{Runtime}\\\textbf{(ms/replan)}} \\
\midrule
Nearest
& \(175.1\pm14.5\)
& \(42.7\pm6.6/122.9\pm19.9\)
& \(183.7\pm16.1\)
& \(328.7\pm29.9\)
& \(8.1\pm10.0/9.3\pm6.8\)
& \(7.23\pm2.10\) \\

Waiting-aware
& \(194.2\pm18.5\)
& \(55.6\pm8.8/172.5\pm35.0\)
& \(202.0\pm22.5\)
& \(414.4\pm50.1\)
& \(14.5\pm9.3/24.3\pm8.3\)
& \(8.55\pm2.60\) \\

MLP--PPO
& \(198.1\pm17.3\)
& \(55.1\pm9.2/173.3\pm40.5\)
& \(206.6\pm24.8\)
& \(421.4\pm45.4\)
& \(16.9\pm8.2/24.5\pm8.5\)
& \(16.54\pm4.26\) \\

Transformer--PPO
& \(200.0\pm19.7\)
& \(59.9\pm8.5/180.3\pm29.7\)
& \(205.9\pm20.9\)
& \(430.8\pm50.5\)
& \(13.9\pm7.0/23.3\pm8.3\)
& \(19.69\pm4.08\) \\

Rolling horizon
& \(216.0\pm20.6\)
& \(34.6\pm4.7/106.3\pm19.3\)
& \(177.3\pm15.5\)
& \(377.4\pm36.2\)
& \(13.6\pm5.1/30.5\pm10.2\)
& \(587.50\pm118.03\) \\
\bottomrule
\end{tabular*}

\vspace{1pt}
\parbox{\textwidth}{
\fontsize{6.2}{6.9}\selectfont
All methods achieved 100\% completion without invalid actions or commitment violations. Slash-separated columns report average/maximum waiting and reassignment/sequence changes.
}
\end{table*}

\subsection{Common-Protocol Comparison of Principal Methods}
\label{subsec:common_protocol}

The five principal methods were evaluated over the same 20 independently generated test scenarios using the common commitment-aware protocol. For each scenario, nearest-feasible assignment, waiting-aware insertion, MLP--PPO, Transformer--PPO, and rolling-horizon optimization received identical request arrivals, customer attributes, vehicle states, depot locations, capacities, and operational events. The same completed, active, and near-term route decisions were protected for every method, and all results were calculated using the objective in \eqref{eq:commitment_combined_objective}.

The previously selected MLP--PPO and Transformer--PPO checkpoints were applied without retraining. The rolling-horizon method optimized the assignment and sequence of the currently flexible requests subject to the same commitment and capacity constraints, with a one-second solution limit at each replanning event. Table~\ref{tab:common_protocol_results} reports the mean and sample standard deviation over the 20 paired scenarios.

All methods completed every request without invalid actions or commitment violations. Nearest feasible achieved the lowest objective, distance, disruption, and runtime. Rolling horizon achieved the lowest waiting and makespan but required greater distance, more sequence changes, and substantially higher computation. MLP--PPO and Transformer--PPO were approximately \(35.5\) and \(29.8\) times faster than rolling horizon, respectively, although the heuristics remained fastest. No method was best across all of these performance criteria.

\section{Conclusion}
\label{sec:conclusion}

This paper presented a stability-aware event-driven framework for dynamic multi-depot routing that combines deterministic feasibility masking, fixed-prefix/flexible-suffix commitments, MLP and Transformer policies trained by BC and PPO, and rolling-horizon benchmarking. Reassignment and resequencing are measured separately to capture replanning disruption.

Across the 20-scenario benchmark, all methods completed every request without invalid assignments. Nearest feasible achieved the lowest mean objective, was best in 18 scenarios, and outperformed the learned policies in distance, waiting, stability, makespan, and runtime. Across five training runs, PPO had little average effect on the MLP and improved the Transformer by approximately \(2.5\%\), but with greater seed variability.

Under the common protocol, nearest feasible again achieved the lowest combined objective and route disruption, whereas rolling horizon achieved the lowest waiting times and makespan at substantially higher computational cost. The learned policies produced feasible routes within tens of milliseconds and transferred from 30 to 80 requests without retraining, but did not outperform the strongest heuristic.

Overall, the results demonstrate the value of integrating feasibility, route commitments, and stability measures into dynamic-routing policies while emphasizing the importance of strong heuristic baselines. Future work will consider multi-size training, improved rewards, time windows, heterogeneous fleets, time-dependent travel, and real transportation networks.

\bibliographystyle{IEEEtran}
\bibliography{VRP_refs}

\end{document}